\documentclass[runningheads]{llncs}
\usepackage{array}
\usepackage{makecell}
\usepackage{multirow}
\usepackage{multicol}
\usepackage{graphicx}
\usepackage{cite}
\usepackage{amsmath,amssymb,amsfonts}
\usepackage{algorithmic}
\usepackage{graphicx}
\usepackage{textcomp}
\usepackage{subfigure}
\usepackage{booktabs} 
\usepackage{caption}
\usepackage{url}
\usepackage{hyperref}
\usepackage{cleveref}
\usepackage{soul}
\usepackage{xcolor}

\newcommand{\samelineand}{\qquad}
\newcommand{\ours}[1]{SDI-GAN}

\begin{document}

\title{Selectively increasing the diversity of GAN-generated samples}

\author{
Jan Dubiński\inst{1} \and
Kamil Deja\inst{1} \and
Sandro Wenzel\inst{2} \and \\
Przemysław Rokita \inst{1} \and
Tomasz Trzcinski\inst{1,3,4}}
\authorrunning{J. Dubiński et al.}

\institute{$^1$Warsaw University of Technology
\samelineand  $^2$CERN \\
\samelineand $^3$Jagiellonian University 
\samelineand $^4$Tooploox \\
\email{jan.dubinski.dokt@pw.edu.pl}}
\maketitle    
\begin{abstract}
Generative Adversarial Networks (GANs) are powerful models able to synthesize data samples closely resembling the distribution of real data, yet the diversity of those generated samples is limited due to the so-called mode collapse phenomenon observed in GANs. 
Especially prone to mode collapse are conditional GANs, which tend to ignore the input noise vector and focus on the conditional information. 
Recent methods proposed to mitigate this limitation increase the diversity of generated samples, yet they reduce the performance of the models when similarity of samples is required.
To address this shortcoming, we propose a novel method to selectively increase the diversity of GAN-generated samples. 
By adding a simple, yet effective regularization to the training loss function we encourage the generator to discover new data modes for inputs related to diverse outputs while generating consistent samples for the remaining ones. More precisely, we maximise the ratio of distances between generated images and input latent vectors scaling the effect according to the diversity of samples for a given conditional input. We show the superiority of our method in a synthetic benchmark as well as a real-life scenario of simulating data from the Zero Degree Calorimeter of ALICE experiment in LHC, CERN.

\keywords{Generative Adversarial Networks, Generative Models, Data Simulation}
\end{abstract}
\section{Introduction}

Generative Adversarial Networks (GANs) \cite{goodfellow2014generative} constitute a gold standard for synthesizing complex data distributions and they are, therefore, widely used across various applications, including data augmentation~\cite{DBLP:journals/corr/abs-2006-03622}, image completion~\cite{DBLP:journals/corr/abs-2103-10428} or representation learning~\cite{zieba2018bingan}. They are also employed in high energy physics experiments at the Large Hadron Collider (LHC) at CERN, where they allow to speed up the process of simulating particle collisions~\cite{paganini2017calogan,deja2020e2esinkhorn,kansal2021particle}. In this context, the generative models are used to generate samples of possible detectors' responses resulting from a collision of particles described with a series of physical parameters. 
For more controllable simulations, we can condition the generative models with additional parameters of the collision, using conditional GANs (cGANs)~\cite{conditionalGAN}. 

Although by conditioning GANs we obtain more context-dependent generations that are closer to the values observed in real experiments, these models are considered more vulnerable to the so-called mode collapse phenomenon~\cite{NIPS2016_8a3363ab} , observed as a tendency to generate a limited number of different outputs per each conditional prior. This, in turn, significantly reduces the effectiveness of employing GANs for particle collision simulations, as alignment of generated samples with the real data distribution is fundamental for drawing correct conclusions from the performed experiments. 


To address the above-mentioned limitations of cGANs, recent methods~\cite{dsganICLR2019, MSGAN, Liu_DivCo} attempt to increase the diversity of generated samples by modifying the associated cost function. However, they do not consider conditioning the diversity on the input conditioning values, assuming a uniform distribution of diversity across all of them. This assumption is rarely observed in practical applications, for instance in particle collision simulations at CERN diversity of generated samples highly depends on the set of conditioning variables. 

In this work, we identify this shortcoming of existing models and propose a simple, yet effective method to selectively increase the diversity of GAN-generated samples, based on conditioning values. In principle, we introduce a regularization method that enforces GANs to follow diversity observed in the original dataset for a given conditional value.
More exactly, we maximise the ratio of distances between latent vectors of generated images and inputs,  scaling the effect accordingly to the diversity of samples corresponding to a given conditional input. Our approach, dubbed~\ours{}, is readily applicable for conditional image synthesis models and does not require any modification of the baseline GAN architecture. 

We evaluate our method on a challenging task of simulating data from the Zero Degree Calorimeter of the ALICE experiment in LHC, CERN. To better demonstrate the performance of our method we also include a synthetic dataset for 2D point generation. We compare our approach with competing methods and achieve superior results across all benchmarks.

The main contribution of this paper is a novel method for increasing the diversity of GAN-generated results for a selected subset of conditional input data while keeping the consistency of the results for the remaining conditional inputs.

\section{Related work}

\subsubsection{Generative simulations:}
The need for simulating complex processes exists across many scientific domains. In recent years, solutions based on generative machine learning models have been proposed as an alternative to existing methods in cosmology~\cite{10.1093/gigascience/giab005} and genetics~\cite{Rodr_guez_2018}.
However, one of the most profound applications for generative simulations is in the field of High Energy Physics, where machine learning models can be used as a resource-efficient alternative to classic Monte Carlo-based\cite{incerti2018geant4} approaches.

Recent attempts ~\cite{paganini2017calogan, kansal2021particle, Erdmann_2019} leverage solutions based on Generative Adversarial Networks~\cite{goodfellow2014generative} or Variational Autoencoders ~\cite{kingma2013auto}. Although those methods offer considerable speed-up of the simulation process, they also suffer from the limitations of existing generative models. Controlling the diversity of simulated results while maintaining the high fidelity of the simulation is one of the challenges of using generative models for such applications. 

\subsubsection{Mode collapse and sample diversity in cGAN:}

The authors of MS-GAN \cite{MSGAN} address the mode collapse problem in cGANs by introducing mode-seeking loss. This additional regularization term added to the generator training function aims to improve generation diversity by maximizing the dissimilarity between two images generated from two different latent codes. During training, the generator tries to minimize the regularization term added to the loss function equal to the inverse of measured diversity.

DS-GAN \cite{dsganICLR2019} tries to tackle the problem with a similar approach. The main difference between the two methods is the fact that DS-GAN explicitly maximizes the measured diversity which is subtracted from the training loss of the generator. 

In DivCo \cite{Liu_DivCo} the authors use contrastive learning to achieve diverse conditional image synthesis. They introduce a latent-augmented contrastive loss which encourages images generated from distant latent codes to be dissimilar and those generated from close latent codes to be similar. The similarity of images is measured using their latent representations extracted from the discriminator network.

Our approach shares a similar method of calculating the diversity of images with \cite{MSGAN} and \cite{dsganICLR2019}. However, contrary to those approaches we do not base our measure of diversity on pixels of generated images. Instead we operate on image representation, similarly to \cite{Liu_DivCo}. Moreover, in principle, all previously described approaches do not account for different levels of variance of samples corresponding to different conditional inputs and instead maximize the diversity of the results generated for all conditional inputs.

\section{Methodology}
In this work we propose a novel selective diversity regularization method that improves alignment between generations from cGAN and data conditioned on available conditional values. For a dataset of images $\mathcal{X}$ with conditioning values $\mathcal{C}$ we want to learn a generator $G$ that is able to produce realistic samples from the domain of $\mathcal{X}$ conditioned on $c \in \mathcal{C}$. Moreover, we want the synthesised images to be diverse or similar to each other depending on the variance of samples in $\mathcal{X}$ that correspond to a given $c$.

Traditional conditional GANs are trained using adversarial loss. Given a condition vector $c \in \mathcal{C}$ and a $k$-dimensional latent code $z \sim \mathcal{N}_{k}(0,1)$ the generator $G$  takes both $c$ and $z$ as input and produces an output image $\hat{x}=G(z, c)$. The image $\hat{x}$ should be indistinguishable from real data by a discriminator $D$. During training generator and discriminator play a min-max game, in which $D$ learns to distinguish real data from samples synthesised by $G$ while $G$ tries to generate samples that are considered by $D$ as real.
\begin{equation}
 \begin{aligned}
\mathcal{L}_{\mathrm{adv}} &(G, D)=\mathbb{E}_{x \sim \mathcal{X}, c \sim \mathcal{C}}[\log D(x, c)] +\mathbb{E}_{c \sim \mathcal{C}, z \sim \mathcal{N}(0,1)}[\log (1-D(G(z, c), c)]
\label{eq:adversarial}
\end{aligned}
\end{equation}

The adversarial loss function encourages the generator to produce realistic data, but as observed by~\cite{Liu_DivCo} it does not directly promote the diversity of synthesised samples.
To alleviate this problem Mao et al. \cite{MSGAN} propose a regularization term that penalizes the low diversity of generated samples. More precisely, the introduced method maximizes the ratio of the distance between two images generated from two different latent codes $ z_1 $, $ z_2 $ and the same conditioning value $c$  with respect to the distance between those latent codes. The proposed regularization term is added to the basic loss function from Eq.~\ref{eq:adversarial} 

\begin{equation}
 \begin{aligned}
\mathcal{L}_{\mathrm{ms}}=\left(\frac{d_{\mathbf{I}}\left(G\left(c, \mathbf{z}_{1}\right), G\left(c, \mathbf{z}_{2}\right)\right)}{d_{\mathbf{z}}\left(\mathbf{z}_{1}, \mathbf{z}_{2}\right)}\right)^{-1}
\end{aligned}
\end{equation}

Although, this approach successfully forces the generator to produce dissimilar examples it does not account for different levels of sample diversity for different conditioning input $c$.  To address this issue we propose a simple yet effective modification of the regularization term.

As a data preprocessing step, for each unique conditioning input  $c \in \mathcal{C}$  we calculate the diversity of samples from the dataset $\mathcal{X}$ that correspond to $c$ denoted as $\mathcal{X}_c$. We base our measure of diversity  $ f_{div}$ on the variance of samples. As denoted in Eq. \ref{f_c}, for each set of images $\mathcal{X}_c$ we sum the standard deviation of pixel values with the same coordiantes for images in $\mathcal{X}_c$: 


\begin{equation}
f_{div}(c) = \sum_{i,j} \sqrt{\frac{\sum_t{(x_{ij}^t-\mu_{ij})^2}}{|X_c|}}
\label{f_c}
\end{equation}
where $i$ and $j$ are the pixel coordinates, $t$ is the index of sample $x \in X_c$ and $\mu_{ij}$ is a mean value for a pixel $ij$ from all samples from $X_c$.
We normalize the obtained values of all sample diversity $f_{div}(c)$ to $<0,1>$.

To account for varying levels of sample diversity for each conditioning input $c$ we multiply the regularization term introduced before by $ f_{div}(c)$. 

\begin{equation}
 \begin{aligned}
\mathcal{L}_{\mathrm{div}}=f_{div}(c)*\left(\frac{d_{\mathbf{I}}\left(G\left(c, \mathbf{z}_{1}\right), G\left(c, \mathbf{z}_{2}\right)\right)}{d_{\mathbf{z}}\left(\mathbf{z}_{1}, \mathbf{z}_{2}\right)}\right)^{-1}
\end{aligned}
\end{equation}

This change forces the generator to better match the diversity observed in the original dataset with respect to conditional values. Intuitively, the generator produces more diverse images under conditions that allow for a higher variety of synthesised samples. At the same time, the generator is not forced to produce dissimilar results if the conditioning value corresponds to a set of similar images in the dataset. The overall objective of training \ours{} is the following: 
\begin{equation}
 \begin{aligned} 
\mathcal{L} = \mathcal{L}_{\mathrm{adv}}(G, D)+\lambda_{\mathrm{div}} \mathcal{L}_{\mathrm{div}}(G)
\end{aligned}
\end{equation}
where $\lambda_{\mathrm{div}}$ is a hyperparameter controlling the strength of the regularization.

Additionally, to better adapt our method to the task of applying cGANs as a fast simulation tool, we propose to base the distance $d_g$ on the dissimilarity of latent representations of images rather than the dissimilarity of pixels. We measure the distance between two generated images by calculating the  L$_1$ metric between their latent representations. We obtain those representations by treating the initial layers of the discriminator as an encoder $E$ and extracting the latent features of the images from its penultimate layer during training.

\begin{equation}
e_{z c} = E(G(z,c))
\end{equation}
\begin{equation}
d_g(G(z_1,c), G(z_1,c)) = |e_{z_1 c},e_{z_2 c}|
\end{equation}

This change shifts the focus of the generator from the visual dissimilarity of images to the difference in their underlying characteristics extracted by the encoder.

\section{Experiments}
We evaluate our method on simulating data from the Zero Degree Calorimeter from the ALICE experiment in LHC, CERN.
Additionally, we use a synthetic dataset as a benchmark. We compare our approach to a conditional DC-GAN \cite{radford2015unsupervised}, MS-GAN \cite{MSGAN} and DivCo \cite{Liu_DivCo}. 

\subsection{Synthetic dataset}
\begin{figure}
\centering
\vspace{-1.5em}
\begin{tabular}{c c c c c}
        \textit{spread} = False & & & &\textit{spread} = True \\ 
        \includegraphics[scale = 0.25]{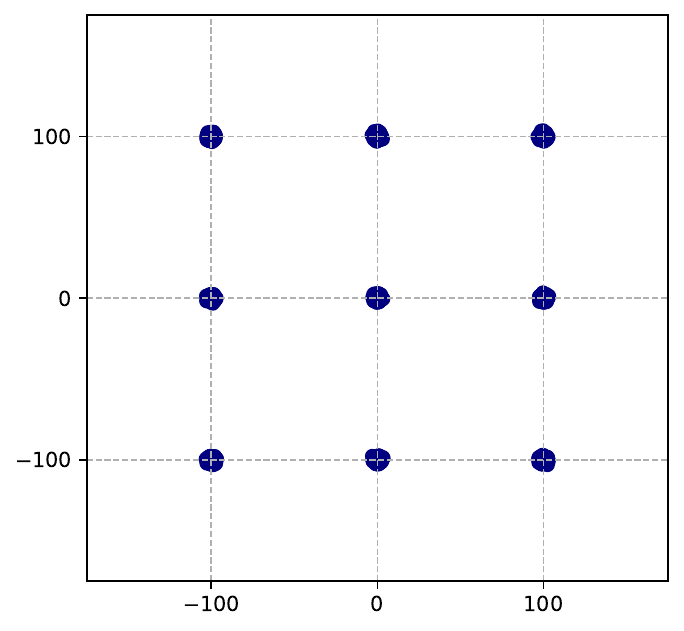} & \makecell{ } & \makecell{ } & \makecell{ } & \includegraphics[scale = 0.25]{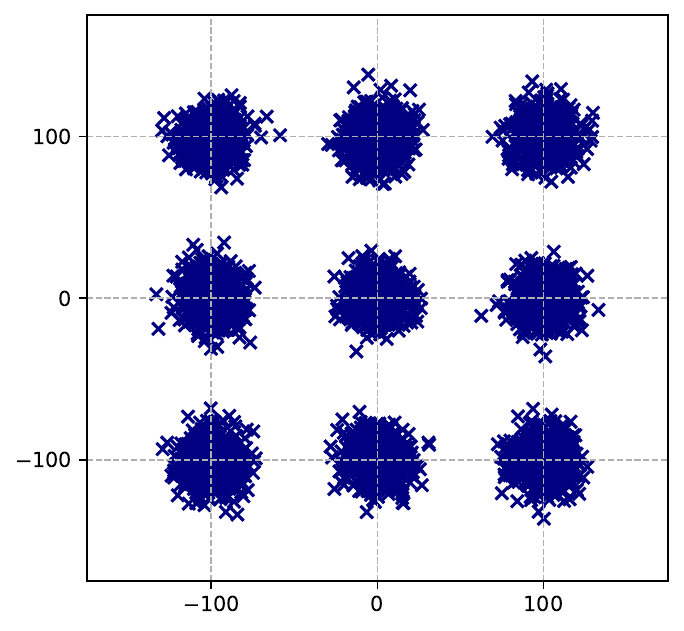}\\
    \end{tabular}
\caption{Synthetic dataset. For each class the generated points form a cluster. For point with \textit{spread} = False the variance of each cluster is equal to 1 and for points with \textit{spread} = True the variance of each cluster is equal to 100.}
\vspace{-30em}
\label{fig:toy_dataset} 
\end{figure}

To clearly demonstrate the effects of our method and its comparison to competing approaches we provide a synthetic dataset of 2D points generation. Each point is conditioned on a class label (from 1 to 9) and a binary variable \textit{spread}. The position of the generated point depends mainly on its class label. However,  points with \textit{spread} = False are generated very close to each other, while points with \textit{spread} = True are heavily dispersed. The dataset is presented in Fig. \ref{fig:toy_dataset}. 

\clearpage

\subsection{Zero Degree Calorimeter simulation}
The task of simulating the response of the Zero Degree Calorimeter (ZDC) offers a challenging benchmark for generative models.
The dataset consists of 295867 samples obtained from the GEANT4 \cite{incerti2018geant4} simulation tool. Each response is created by a single particle described with 9 attributes (mass, energy, charge, momenta, primary vertex).

During the simulation process, the particle is propagated through the detector for over 100 meters while simulation tools must account for all of its interactions with the detector’s matter. The end result of the simulation is the energy deposited in the calorimeter’s fibres, which are arranged in a grid with 44 × 44 size. We treat the calorimeter’s response as a 1-channel image with 44 × 44 pixels, where pixel values are the number of photons deposited in a given fibre. To create the dataset the simulation was run multiple times for the same input particles. For that reason, multiple possible outcomes correspond to the same particle properties.  We refer to this dataset as HEP.

\begin{figure}[h!]
\centering
\begin{tabular}{c c c}
        &  \makecell{diverse\\results}  & \makecell{\includegraphics[scale = 0.64]{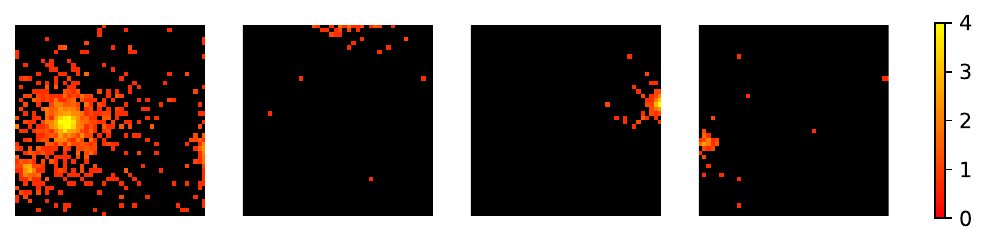}}\\
        & \makecell{consistent\\results}  & \makecell{\includegraphics[scale = 0.64]{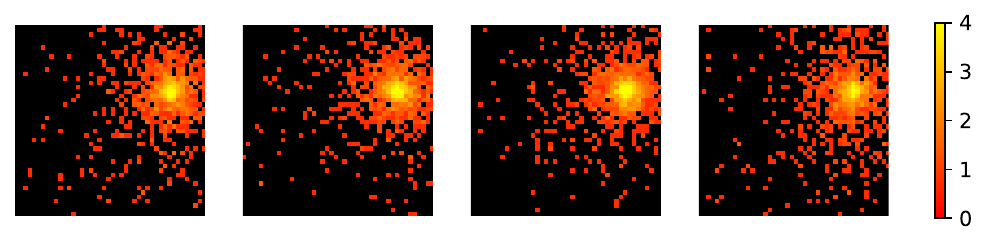}}\\
    \end{tabular}
\caption{Examples of ZDC calorimeter responses. We show 4 possible outputs of the simulation generated for 2 distinct particles.}
\label{fig:HEP_dataset} 
\end{figure}

Although the process that governs the propagation of the particles is non-deterministic by nature, the majority of particles create consistent ZDC responses. However, a subset of particles produces highly diverse results and allows for multiple possible calorimeter responses. In Fig.~\ref{fig:HEP_dataset} we present sampled simulations for two different conditional values. In the first row, we depict the calorimeter response for a high-energy proton, while for the second the input particle is a neutron, which has no electric charge, therefore responses observed in the calorimeter are much more consistent.

\clearpage

\subsection{Results}

\begin{table}[t]
\def\arraystretch{1.1}
\setlength\tabcolsep{0.2cm}
  \caption{Results comparison on the synthetic and HEP datasets. Our solution outperforms competing approaches by generating data with variance close to the real data for both types of conditioning input in the synthetic case. We achieve the lowest Wasserstein distance between test and generated coordinates for the diverse points without any significant trade-off for the consistent points.   The performance improvement introduced by \ours{} is further evaluated on the real-life HEP dataset, where \ours{} achieves the lowest Wasserstein distance between channels calculated from original and generated data.}
  \label{tab:all_results}
  \centering
  \begin{tabular}{l|cc|cc|c}
    \toprule
    &\multicolumn{4}{c|}{Synthetic dataset} & HEP             \\  
    \cmidrule(r){1-6}
    &  \multicolumn{2}{c|}{Variance}  &  \multicolumn{2}{c|}{Wassesrstein
    $\downarrow$} & Wassesrstein
    $\downarrow$\\
    \textit{Spread} &  0 & 1  &  0 & 1 & - \\
    \midrule
    Real & 1 & 100 & - & - & - \\
    DC-GAN & \textbf{0.2} & 2.3 & \textbf{0.7} & 7.4 & 7.6 \\
    MS-GAN & 115.5 & 280,7  & 9.0 & 6.7 &21.7 \\
    DivCo & 163.3 & 3.7 & 1.1 & 7.0 & 14.3 \\
    \textbf{\ours{}} \textbf{(ours)}& 3.3 & \textbf{127.4} & 1.2 & \textbf{2.1} & \textbf{4.5}  \\
    \bottomrule
  \end{tabular}
\end{table}

\begin{figure}[h!] 
\centering
\begin{tabular}{m{1em}  c c c c c}
        & real data & \makecell{\textbf{\ours{}} \\ \textbf{(ours)}} & DC-GAN & MS-GAN & DivCo\\
        \rotatebox{90}{\textit{spread}= 1}& \makecell{\includegraphics[scale = 0.18]{figures/real_toy_diverse.pdf}} & \makecell{\includegraphics[scale =  0.18]{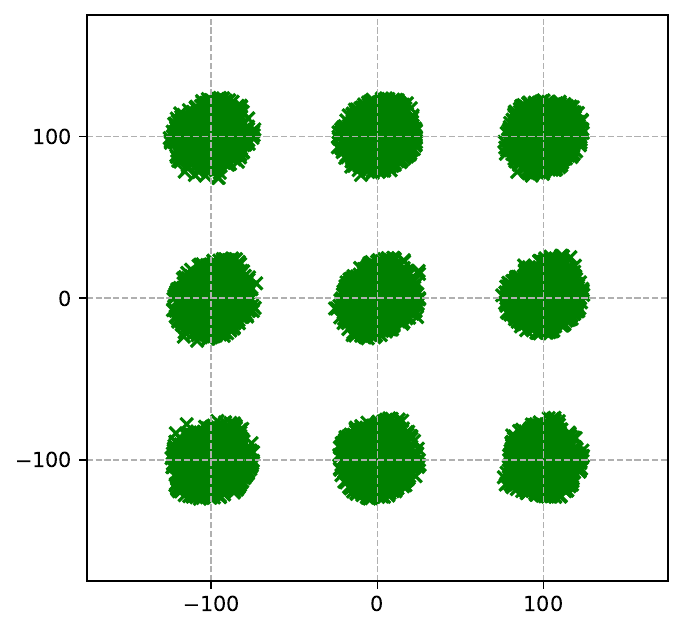}} & \makecell{\includegraphics[scale =  0.18]{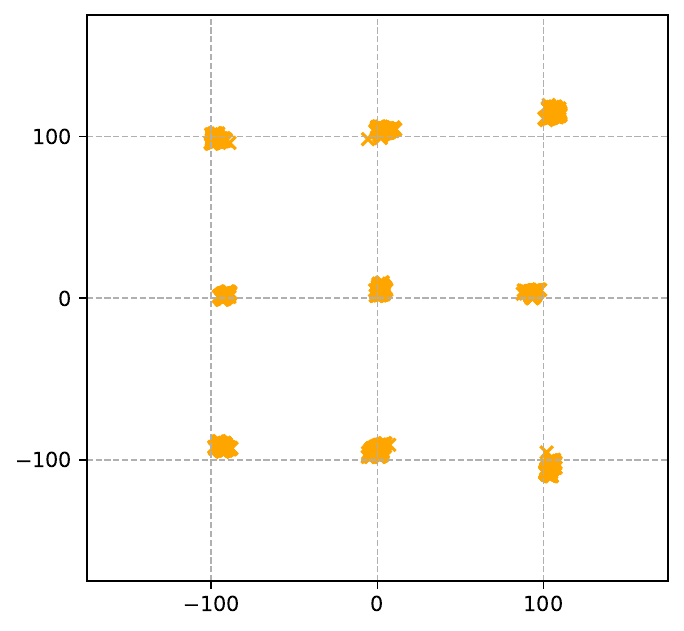}} & \makecell{\includegraphics[scale =  0.18]{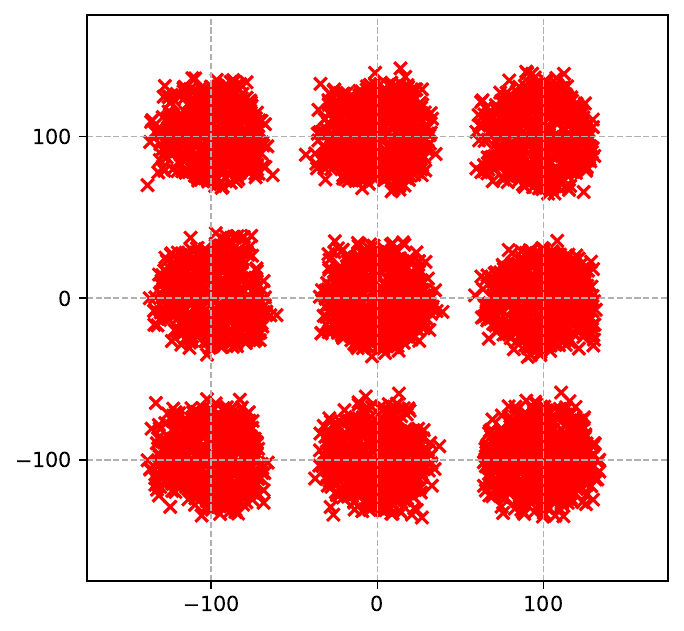}}& \makecell{\includegraphics[scale =  0.18]{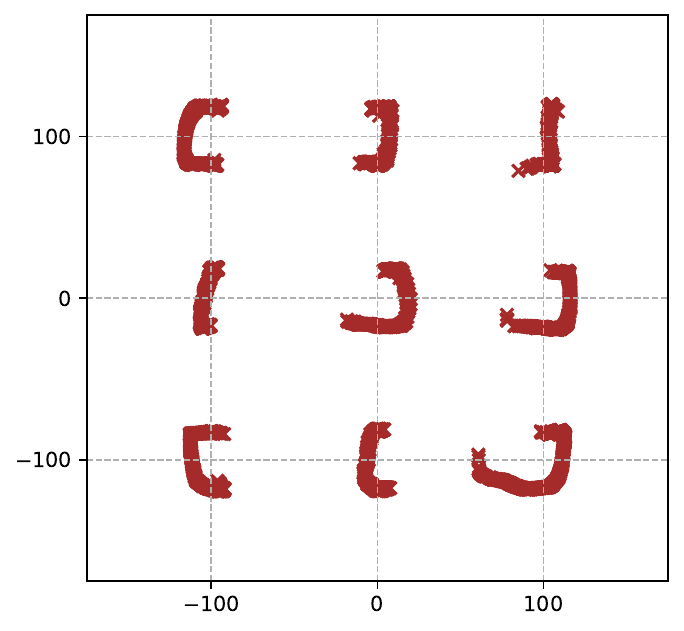}}\\
        \rotatebox{90}{\textit{spread} = 0}& \makecell{\includegraphics[scale =  0.18]{figures/real_toy_consistent.pdf}} & \makecell{\includegraphics[scale = 0.18]{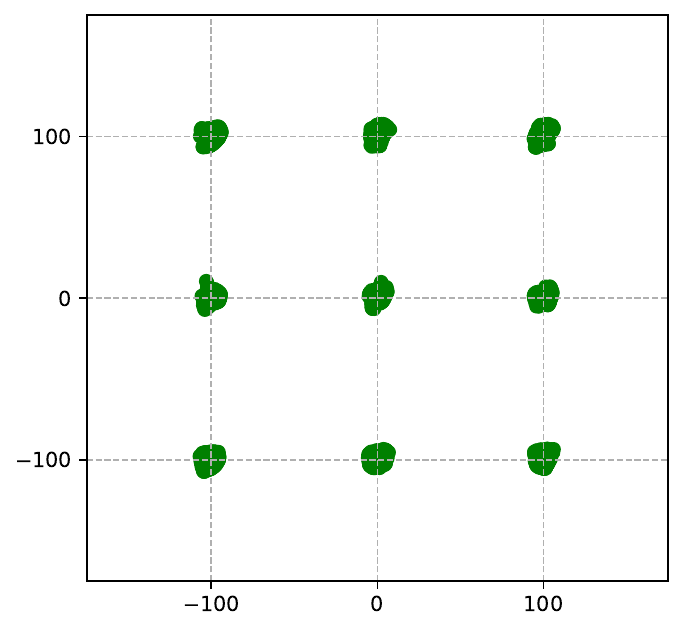}} & \makecell{\includegraphics[scale =  0.18]{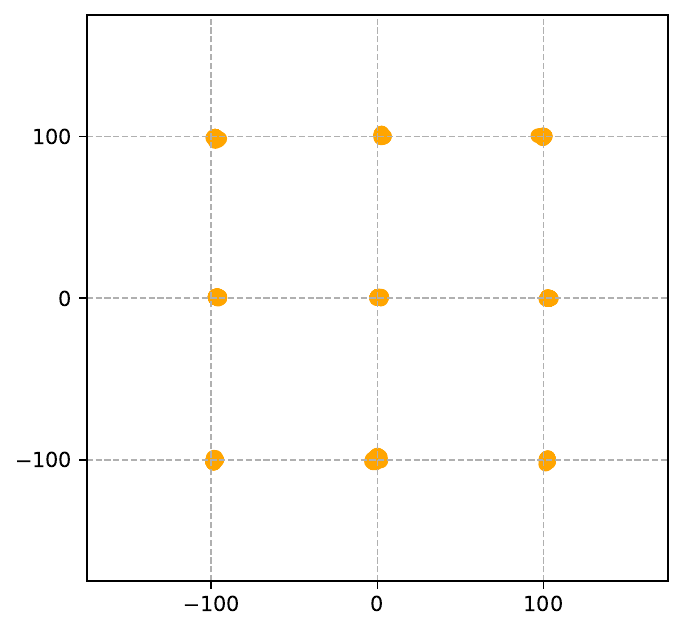}} & \makecell{\includegraphics[scale =  0.18]{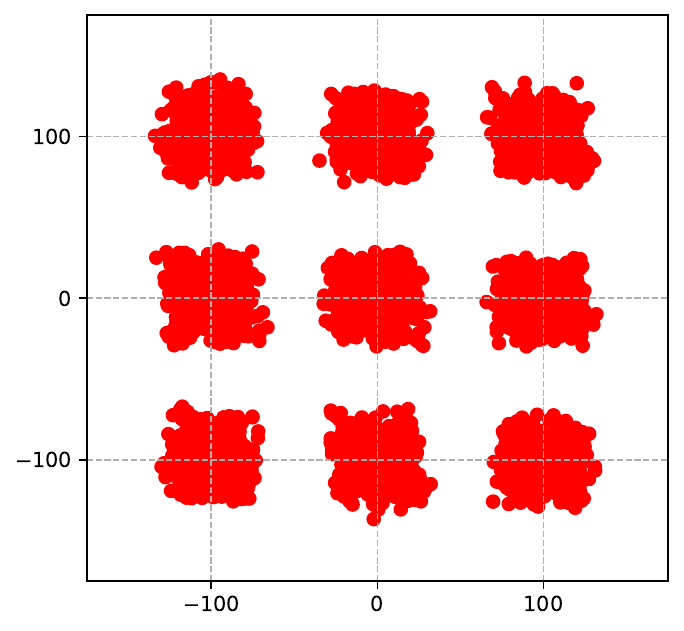}}& \makecell{\includegraphics[scale =  0.18]{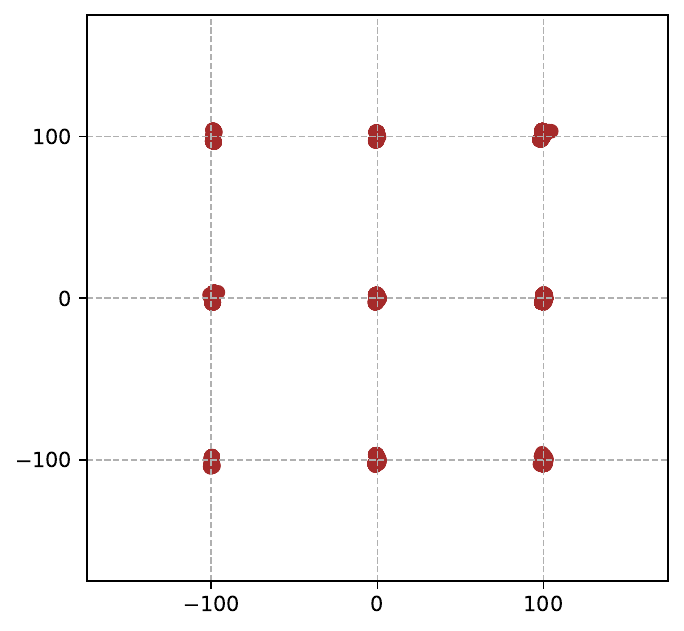}}\\
    \end{tabular}
\caption{Comparison of generated results for the synthetic dataset. Contrary to competing methods, our approach is able to properly synthesise both diverse and similar samples depending on the conditioning variable \textit{spread}.}
\vspace{0.5cm}
\label{fig:toy_results} 
\end{figure}

We present the results of our experiments on the synthetic dataset using both qualitative and quantitative comparisons. As shown in Fig. \ref{fig:toy_results} our method generates samples that are visually most similar to the real data. To confirm this observation we calculate the mean variance for all classes for points with \textit{spread = 0} and \textit{spread = 1}. The results in Tab. \ref{tab:all_results} show the superiority of our method in this scenario. DC-GAN is able to produce results with variance close to real data for conditional inputs with \textit{spread} = 0, but fails to generate diverse results. MS-GAN properly reflect the diversity of possible results for conditional inputs with \textit{spread} = 1, but does not generate consistent results. Although samples created by DivCo have different variance depending on the conditioning input, the data distribution generated under condition \textit{spread} = 1 is distorted and does not match the real data, as presented in Fig.~\ref{fig:toy_results}. Our approach is able to produce both diverse and similar results depending on the conditional information.

The most common method for evaluating GANs on real datasets utilizes Frechet Inception Distance (FID) \cite{heusel2017gans}. However, for the HEP dataset, we propose a domain-specific evaluation scheme that better measures the quality of the simulation. Following the calorimeter’s specification \cite{dellacasa1999alice} we base our evaluation procedure on 5 channels calculated from the pixels of generated images. 
These channels reflect the physical properties of simulated collision and are used for analysing the output of the calorimeter.
To measure the quality of the simulation we compare the distribution of channels for the original and generated data using Wasserstein distance \cite{tolstikhin2017wasserstein}.

\begin{figure}[h!]
\centering
\begin{tabular}{c c c }
     &\textbf{diverse results} & \textbf{consistent results} \\
     real data & \makecell{\includegraphics[scale = 0.4]{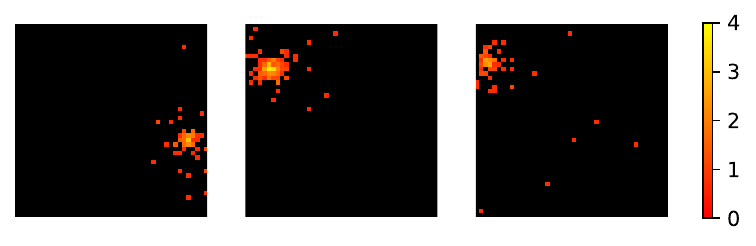}} & \makecell{\includegraphics[scale = 0.4]{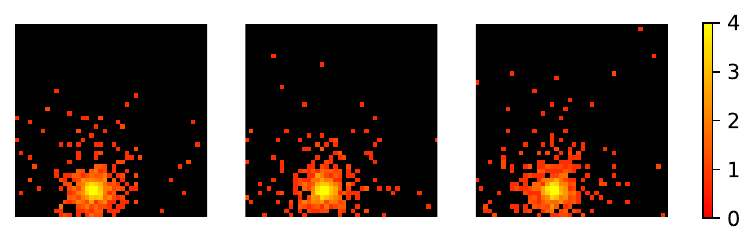}}\\
     DC-GAN & \makecell{\includegraphics[scale = 0.4]{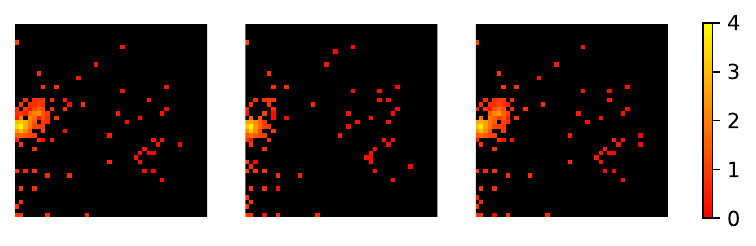}} & \makecell{\includegraphics[scale = 0.4]{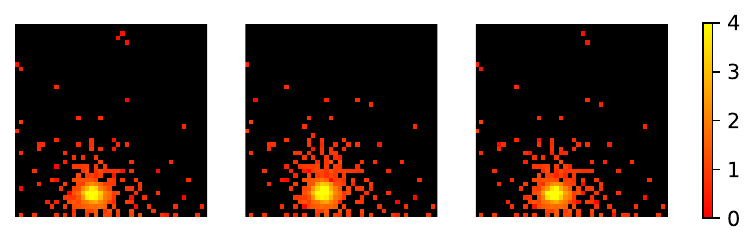}}\\
     MS-GAN & \makecell{\includegraphics[scale = 0.4]{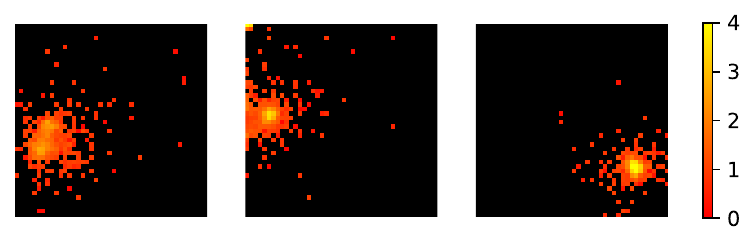}} & \makecell{\includegraphics[scale = 0.4]{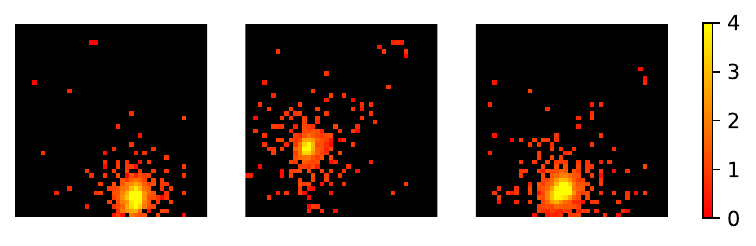}}\\
     DivCo & \makecell{\includegraphics[scale = 0.4]{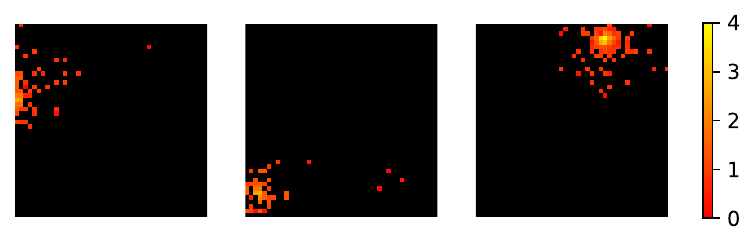}} & \makecell{\includegraphics[scale = 0.4]{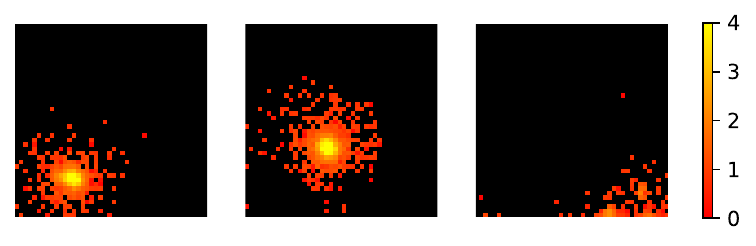}}\\
      \makecell{\textbf{\ours{}} \\ \textbf{(ours)}} & \makecell{\includegraphics[scale = 0.4]{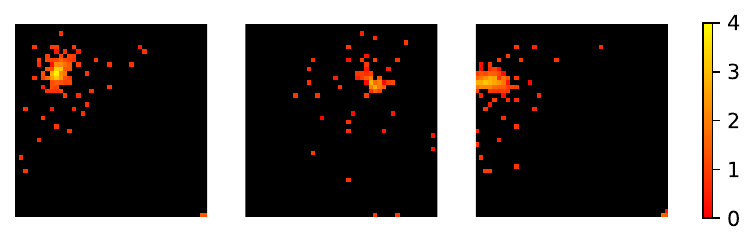}} & \makecell{\includegraphics[scale = 0.4]{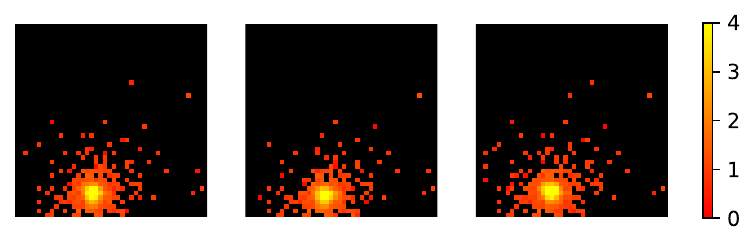}}\\
          
    \end{tabular}
\caption{ Examples of calorimeters response simulations with different methods. DC-GAN works well for particles with consistent responses but fails to generate diverse outcomes when needed. Although MS-GAN and DivCo successfully increase the diversity of generated samples those models do not distinguish between particles that should produce diverse or consistent showers. Our method is able to generate diverse results while producing consistent responses for appropriate particles.
} 
\vspace{-3em}
\label{fig:HEP_examples}
\end{figure} 

\clearpage

\begin{figure}[h!]
\centering
\begin{tabular}{c c c }
     &\textbf{Channel 1} & \textbf{Channel 2} \\
     DC-GAN & \makecell{\includegraphics[scale = 0.18]{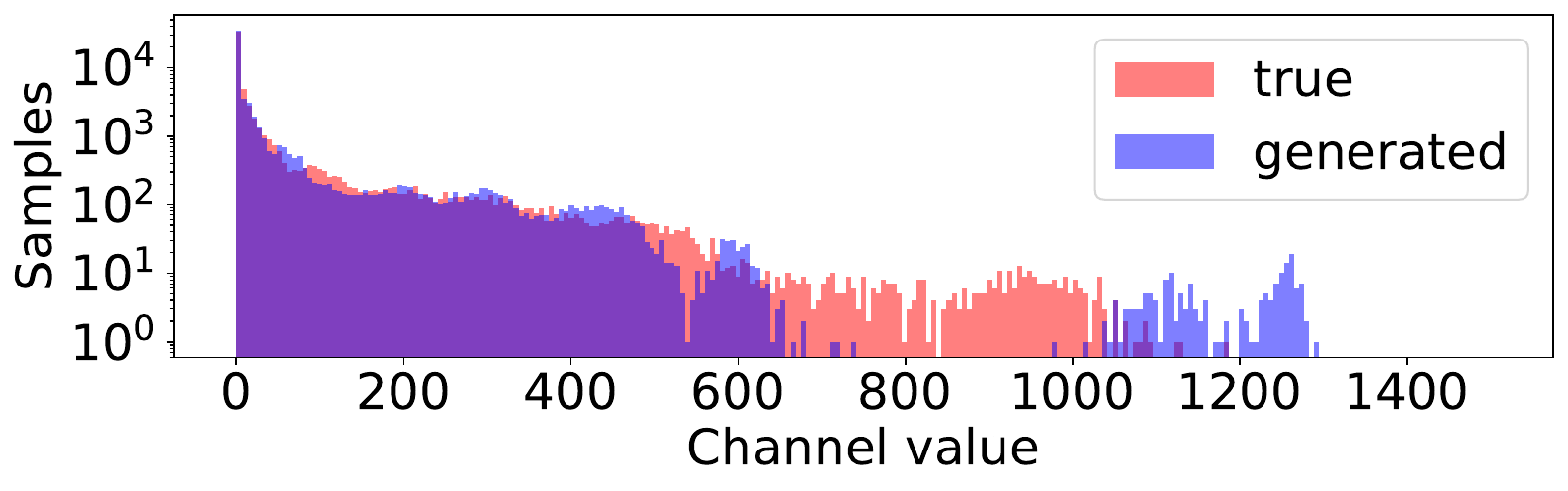}} & \makecell{\includegraphics[scale = 0.18]{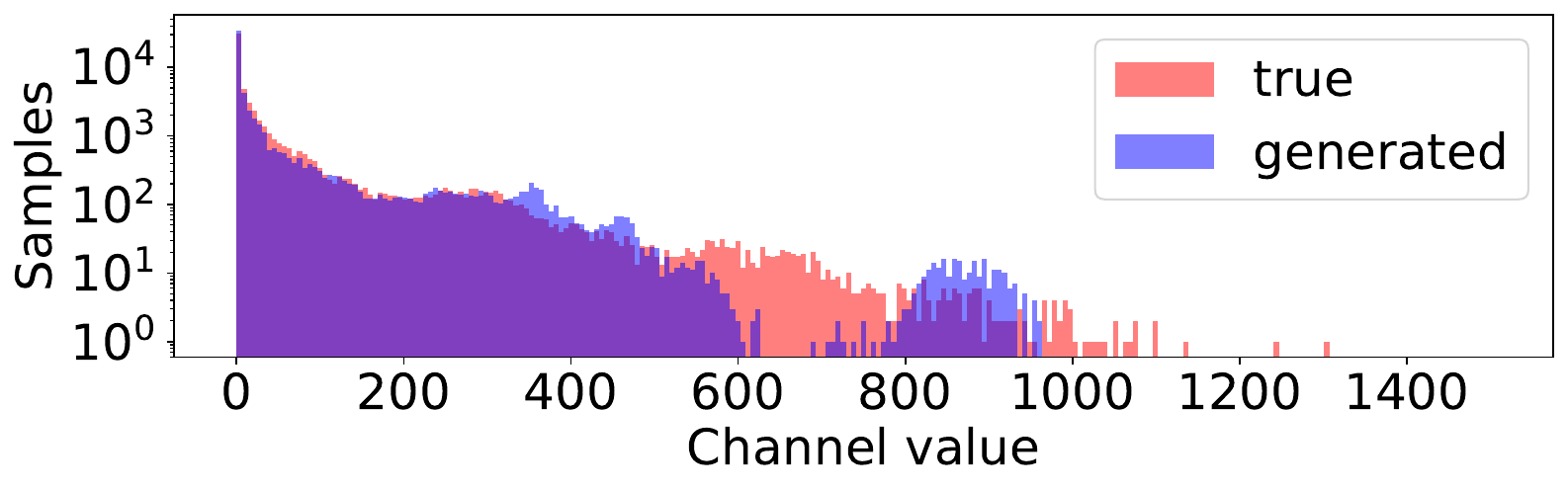}}\\
     MS-GAN & \makecell{\includegraphics[scale = 0.18]{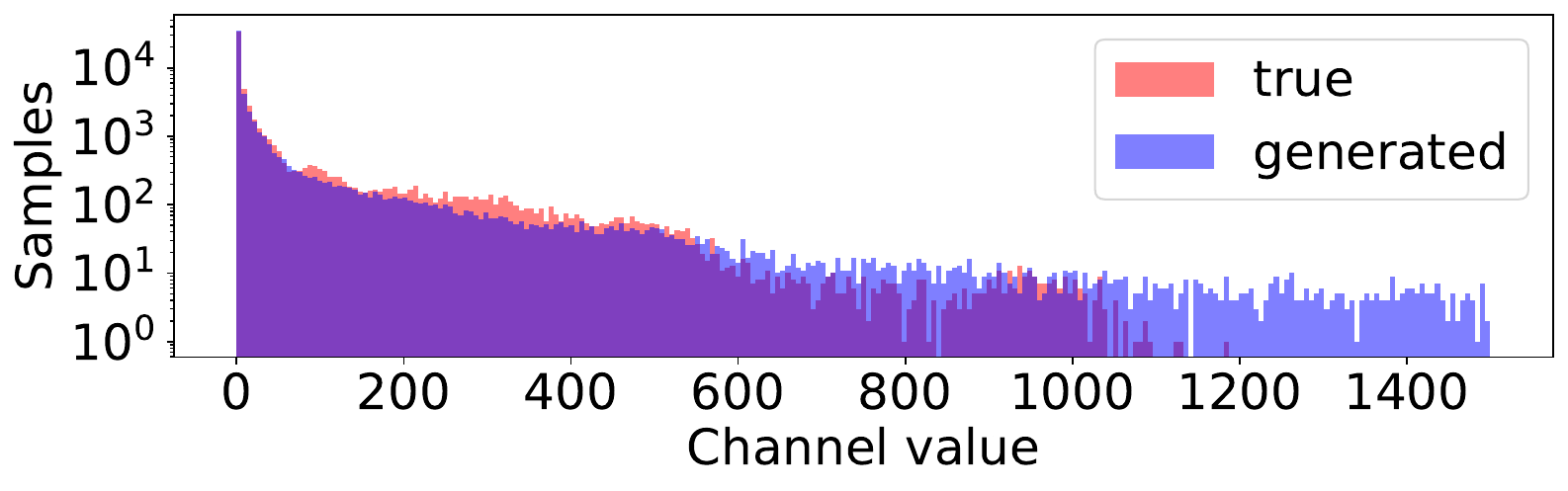}} & \makecell{\includegraphics[scale = 0.18]{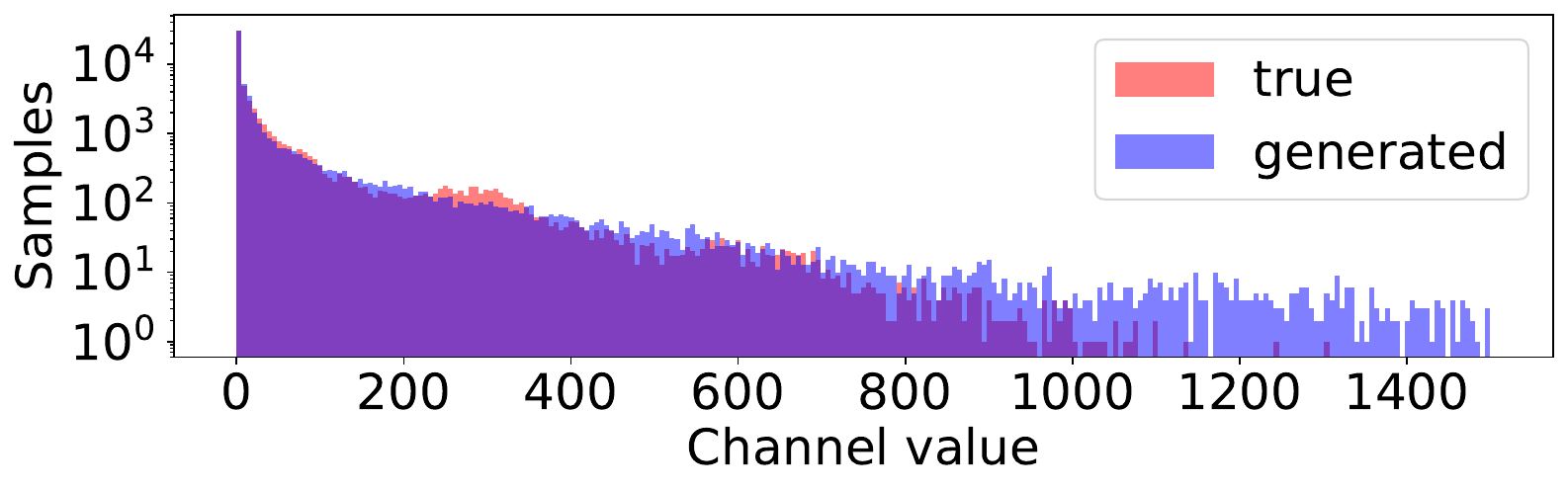}}\\
     DivCo & \makecell{\includegraphics[scale = 0.18]{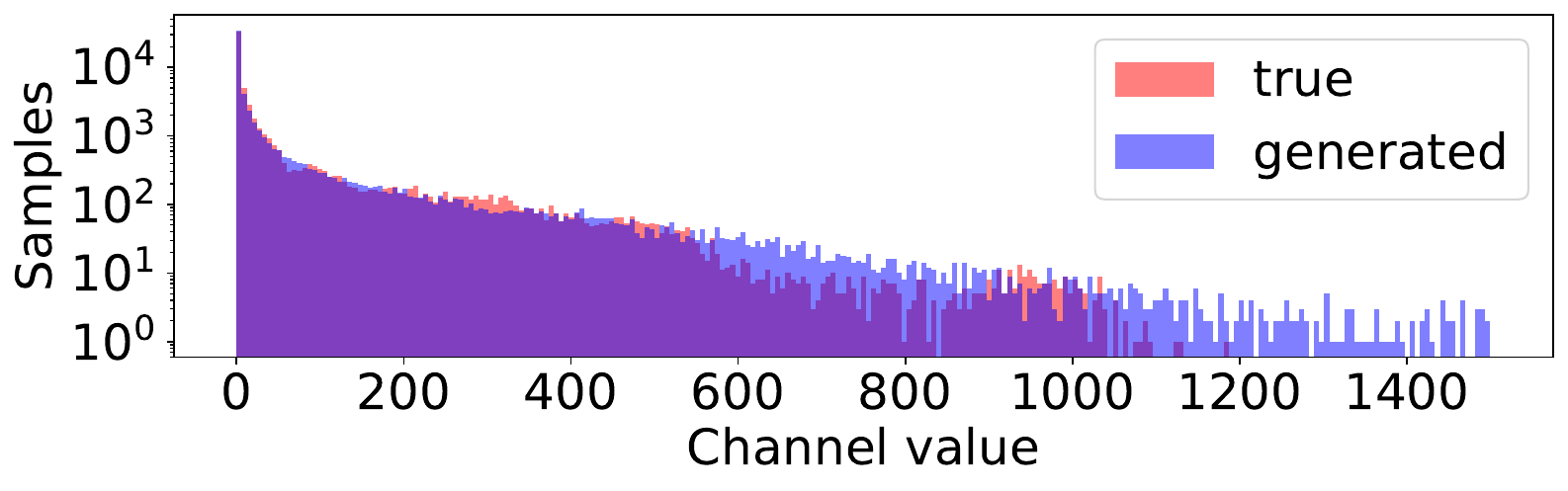}} & \makecell{\includegraphics[scale = 0.18]{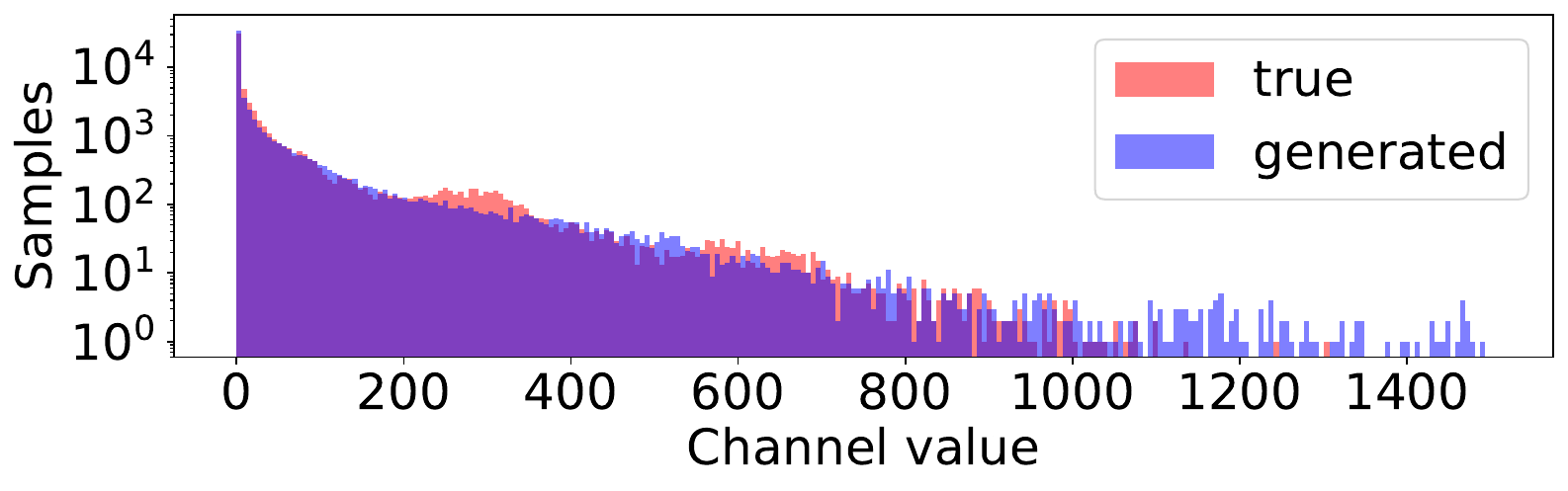}}\\
      \makecell{\textbf{\ours{}} \\ \textbf{(ours)}} & \makecell{\includegraphics[scale = 0.18]{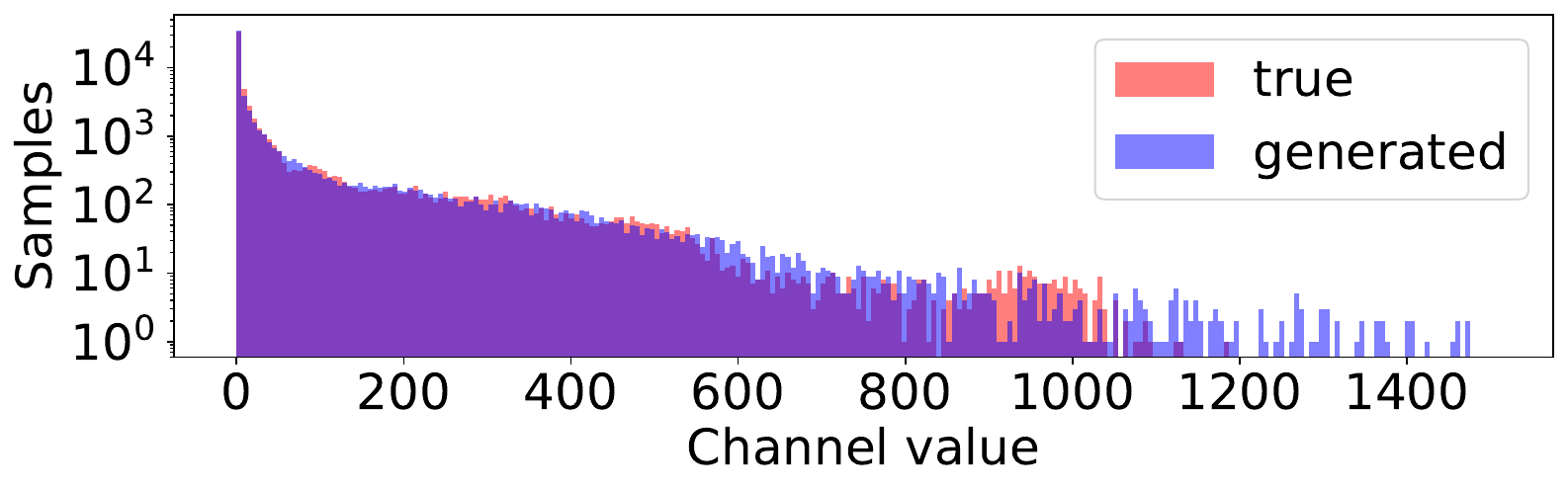}} & \makecell{\includegraphics[scale = 0.18]{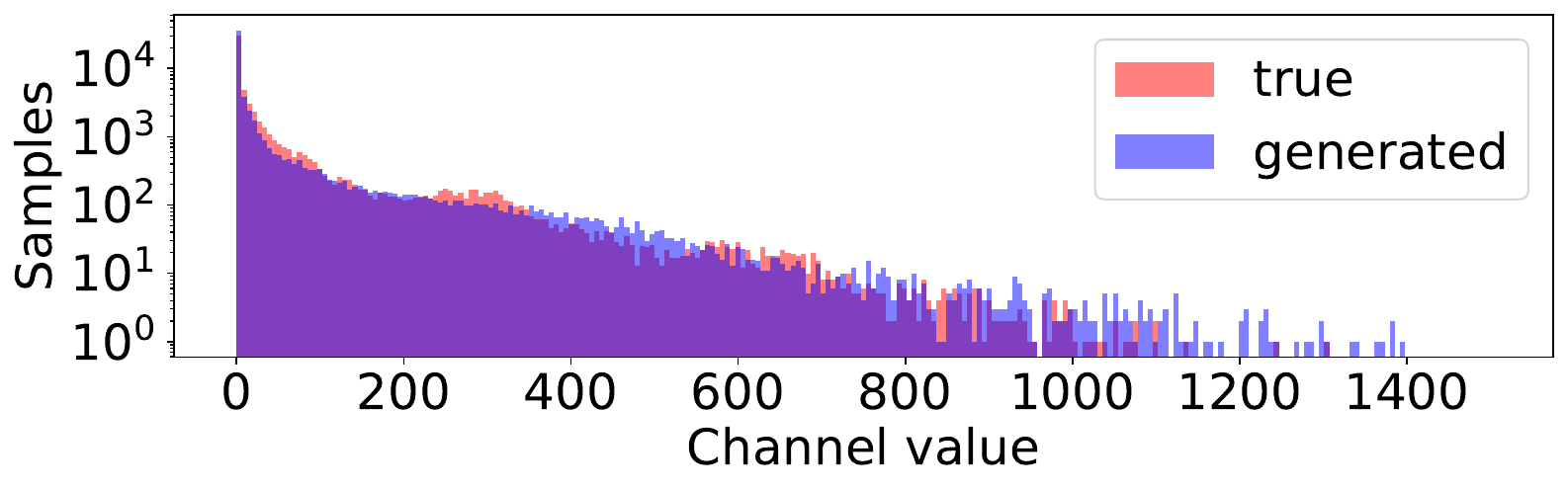}}\\
          
    \end{tabular}
\caption{ Comparison of channel values distribution for selected channels. Our method decreases the differences between the distribution of original and generated data and smooths the distribution of the synthesised results. In the case of \ours{} the increased diversity of generated samples does not harm the fidelity of the simulation, contrary to competing approaches. 
} 
\vspace{-1em}
\label{fig:ws_distributions}
\end{figure}

As presented in Tab.~\ref{tab:all_results}, our approach outperforms other solutions on the HEP datasets. In Fig.~\ref{fig:HEP_examples} we demonstrate that our method is able to generate diverse results for a specific subset of particles while keeping consistent responses for the remaining conditional inputs. The positive impact of this approach on the distribution of the generated samples is further confirmed by Fig.~\ref{fig:ws_distributions} where we compare channel distribution for \ours{} and competing approaches for 2 selected channels. Our method increases the fidelity of the simulation by smoothing the distribution of generated responses and covering the whole range of possible outputs.

The additional regularization term for training of \ours{} does not influence the inference speed of the model. In our initial experiments, we observe a speed-up of simulations of two orders of magnitude when compared to the standard Monte-Carlo approach. With \ours{} this computation boost is observed without degradation in simulation quality. We leave the detailed analysis of this performance gain and the influence of fast simulations on physical experiments for future work. 

\section{Conclusions}
In this work we introduce a simple, yet effective modification of the loss function for conditional generative adversarial networks. 
Our solution enforces increased sample diversity for a subset of conditional data without affecting samples that are characterised by conditional values associated with consistent responses.  

We show that our solution outperforms other comparable approaches on the synthetic benchmark and the challenging practical dataset of calorimeter response simulations in the ALICE experiment at CERN. 

\section*{Acknowledgments}
This research was funded by National Science Centre, Poland grants: no 2020/39 /O/ST6/01478, no 2018/31/N/ST6/02374 and no  2020/39/B/ST6/01511.

\bibliographystyle{splncs04}
\bibliography{mybibliography}
\end{document}